# Reflexivity in Issues of Scale and Representation in a Digital Humanities Project

Annie T. Chen

University of Washington

Camille Lyans Cole

University of Cambridge

**ABSTRACT**

In this paper, we explore issues that we have encountered in developing a pipeline that combines natural language processing with data analysis and visualization techniques. The characteristics of the corpus – being comprised of diaries of a single person spanning several decades – present both conceptual challenges in terms of issues of representation, and affordances as a source for historical research. We consider these issues in a team context with a particular focus on the generation and interpretation of visualizations.

## 1 INTRODUCTION

Digital humanists are often faced with questions from other humanists about the value of digital over more traditional modes of analysis [1]. One area where digital methods have been held up as most useful is in dealing with "big data" [2]. The question of "bigness" in data is not only about scale, but also about "distance" from texts and the interconnectedness of data [3]–[5]. Recently, advocates of "big data" and "distant reading" have faced calls for increased reflexivity and attention to context [6], [7].

Visualization techniques common to digital humanities research – including heat maps, tag clouds, maps, and network graphs – can often contribute to feelings of bigness and distance from texts [4], [6], [8], [9]. In this paper, we highlight how any kind of data, "big" or not, generates questions of scale, while examining how visualization can support analysis at multiple scales. By considering the advantages and disadvantages offered by different scales of visualization and interpretation, we demonstrate the importance of reflexivity, drawing out the role of the research team in shaping the representation of data about people and communities, and focusing on how this affects visualization and analysis.

In recent years, there has been much discussion of distant reading, which involves the generation of an abstract view of the features of a text, e.g., quantitative or qualitative statistics about the text, relationships between the textual elements, or metadata about the textual entities, to facilitate exploration of research questions [4][8]. In his work, *Graphs, Maps, and Trees,* Moretti argues for 'distance' as an end goal, positing that in the reduction of elements, we are able to sharpen the acuity of what is perceived [5]. However, this idea has also raised the concern that in abstraction we will lose sight of context, potentially resulting in the danger of interpreting figures and tables that seemingly support our own arguments [6].

In response, others have argued for combining distant with close reading in interactive visualization [6][8][9]. Jänicke et al.'s review of close and distant reading approaches highlights the prevalence of top-down approaches, which involve starting with a high-level visualization and drilling down to details as needed, and the use of top-down and bottom-up approaches together, which involves employing one visualization system and switching between close and distant reading while studying differences in the views generated [8].

One common distant reading approach has involved network analysis, which can facilitate exploration of connections that exist between people [10]. Visualization of social relations can be used to explore research questions. In historical research, scholars are increasingly using linked data and graph-based approaches to visualize data from a large volume of sources, including biographical dictionaries and digitized historical corpora [10], [11]. In literary studies, analyses can often be focused on a single, lengthier work, which may exhibit great complexity due to the number of persons, or characters, mentioned [12]. These analyses are often based on co-occurrence networks, and sometimes model interactions between characters [13], [14]. In this paper, we describe a non-fictional corpus, but we draw upon network analysis research of literary works because the challenges posed by the diary corpus that we work with resemble many of the challenges encountered in lengthier fictional works. One such example is the coreference resolution of characters appearing in a work who can go unnamed throughout the entire work [14], [15].

Recent work in developing network visualization systems to facilitate exploration of large historical document collections has involved engaging users who were domain experts without a strong background in network science to understand requirements for designing systems [11]. The resulting prototype Intergraph, an interactive system that facilitates exploration of network visualizations based on multilayer graphs, was well-received but also reported challenges. For example, though the developers sought to minimize challenges for users having to learn new software, they reported that users still required significant time to learn how to operate the prototype and to adjust previously established research workflows. In addition, there was a continued need to understand how to adapt automated data extraction and interactive visualization-based exploration to new domains [11].

In this work, we tackled a problem that involved similar challenges as well as new ones. We describe our work developing a pipeline for exploring person entities for use by an interdisciplinary team, including members from differing intellectual traditions and with varying levels of technical ability. We explore the question of how to facilitate discovery and sense-making by considering issues having to do with data representation and interaction modality, and considering how different actors interact with and understand data representations.

* atchen@uw.edu, camillecole@gmail.com

## 2 PROJECT OVERVIEW

### 2.1 The Svoboda Diaries Project

The Svoboda Diaries Project focuses on the preservation of personal diaries written at the turn of the 19th century, which serve as a rich source on 40 years of life, politics, and landscape in Ottoman Iraq. The collection is comprised of two main parts, the Joseph Mathia Svoboda diaries, and the travel journals of Alexander Svoboda, Joseph's son. Of the former, there are 61 volumes known to be in existence. To date, we have digitized three, and on average, the diaries are 375 pages in length.

The diaries of Joseph Mathia Svoboda detail his daily life and regular journeys as a steamboat purser during the late 19th and early 20th centuries, specifically between the cities of Basra and Baghdad. These diaries offer a unique perspective on daily life, community structure, and social relations, in a time and place for which few firsthand accounts of everyday life exist.

As we transcribe this corpus and make the transcriptions available for public use, we also seek to use it to explore our own research questions. The capture and visualization of context is perhaps particularly important for diaries and other forms of autobiographical writing, which can be a rich source of information about everyday life and 'ordinary' acts [16]. Questions of interest to us include how people lived at the time, their social interactions, and temporal change (e.g., seasonal changes in their everyday lives).

### 2.2 Natural Language Processing and Visualization Pipeline

The long-term goal of this project is to develop a natural language processing and visualization infrastructure that could support exploration of a variety of research questions. Our current work has arisen partially out of our planning for the diverse future needs of the project, as well as the human resources that have been involved in the project and will be in the future. The team is comprised largely of undergraduates from many disciplines, as well as an international set of collaborators, and exhibits a fluid structure. This fluid structure is an important consideration, as on the one hand it allows for the introduction of new ideas, enthusiasm, and creativity, and on the other, presents challenges in terms of consistency of the work and data produced.

Consistent with the trend towards open data, open software, and reproducibility in the biomedical sciences [17] as well as in digital humanities research [18], we have developed a pipeline using open-source software, with the code and data hosted in a cloud repository to facilitate geographically distributed collaboration.

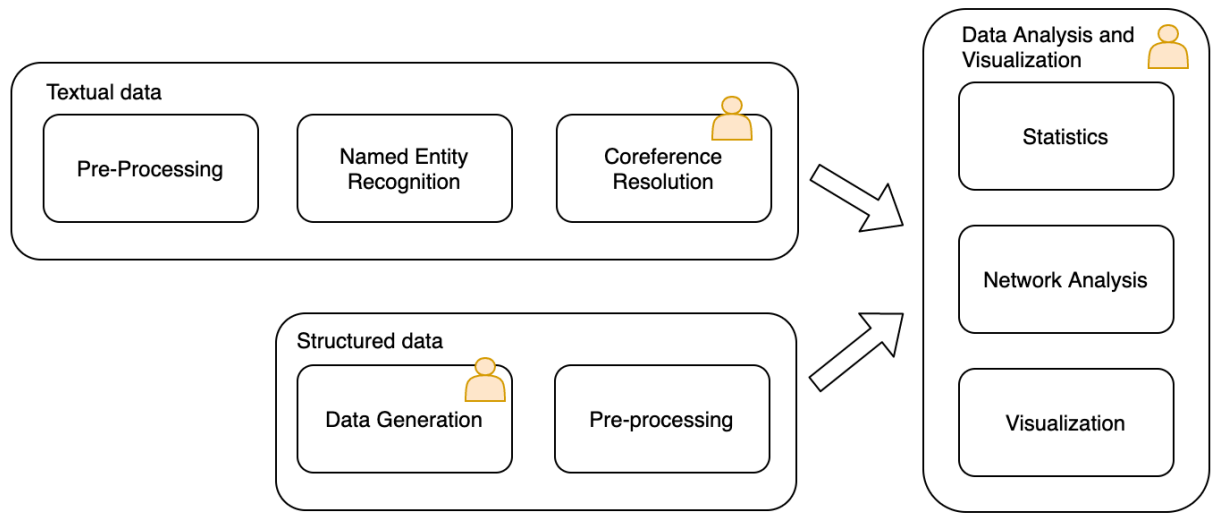

Figure 1: NLP + Data Visualization Pipeline. Person icons represent steps that involve expert curation and/or interpretation.

There are three main parts of this pipeline (Figure 1). The textual data – diaries – are processed and prepared for analysis using NLP techniques. To perform coreference resolution, in which different names referring to the same person are mapped to a single name, we use a hybrid system of coreference resolution that involves rule-based heuristics, dictionary mapping, and human-in-the-loop resolution of extracted entities. The mapping system includes gazetteers and dictionaries containing aliases that are produced iteratively by members of our team of different backgrounds, including social scientists, humanities scholars, and students. We also analyze structured, or tabular, data contributed by humanities scholars that has been produced using close reading and manual examination.

The pipeline has primarily been developed in Python, but we aim to preserve flexibility in the use of different packages and technologies, as they each have affordances and drawbacks depending on the aim. For example, with network analysis we find Gephi to be the most convenient for exploration and discovery, but recognize that other packages such as igraph and networkx can perhaps facilitate some analyses, such as the calculation of metrics on network permutations, more readily. We employ matplotlib, D3, and plotly for other visualizations. Our typical workflow involves pre-processing and transforming data to be analyzed through statistical analysis and visualization. The output of the analytic phases can be adjusted and re-run interactively as researchers manually examine the text. Incorporation of domain experts in visual analytics pipelines is common in information visualization pipelines [19], including ours. The user-centric steps, in which domain experts play a central role in semantic enrichment, results validation, and interpretation, are indicated with person icons (Figure 1).

## 3 REFLEXIVITY IN PRACTICE

Reflexivity is fundamental to our research practice. Reflexivity involves a critical awareness of the role of the researcher and how their backgrounds, assumptions, and perspective may affect the research process [20]. With respect to our data, this is particularly important in terms of awareness of how our decisions shape data transformation, representation, and analysis. Questions also arise about how to use visualizations in the research process.

### 3.1 Considering Network Scale

In his occupation as a steamship purser, Joseph Svoboda encountered a large number of persons in his daily life. Studying persons and how Joseph Svoboda refers to them serves a two-fold purpose of helping us to better understand his social interactions, and offering insight into how different persons might have interacted with one another and been perceived at the time.

The text of the diaries contains many clues that provide some insight into these questions. Depending on the diary, the length of time and number of persons mentioned differ, but the range is approximately 7-9 months, with on average, 7-8 persons (SD~5-6 persons) being mentioned a day, or somewhere in the range of 350-500 persons total in a diary.

In rendering visualizations of Svoboda's social interactions, we have made critical decisions about the number and "nature" of persons to include in network visualizations. For example, consider the use of subcommunity detection in conjunction with network analysis to explore commonalities and differences that may exist in the persons that are mentioned in Svoboda's diaries. By exploring who Svoboda mentioned together in the diaries, we can better understand who he interacted with, in which groups, as well as how those groups might have interacted with one another.

We conceptualized the network as a set of nodes comprised of all persons mentioned in a given diary, with two people being connected by an edge if they were mentioned on the same day. We visualized the networks using Gephi [21], with the *ForceAtlas2* layout [22], a network layout algorithm that leverages energy attraction and repulsion in the network to approximate relation strength, and thus closer social interaction, between person nodes. We employed the LabelAdjust algorithm, which adjusts the spacing between nodes to make the labels (person names) more readable. To identify subcommunities or persons who might share similarities with one another, we employed the Louvain method [23] and colorized the nodes by modularity grouping.

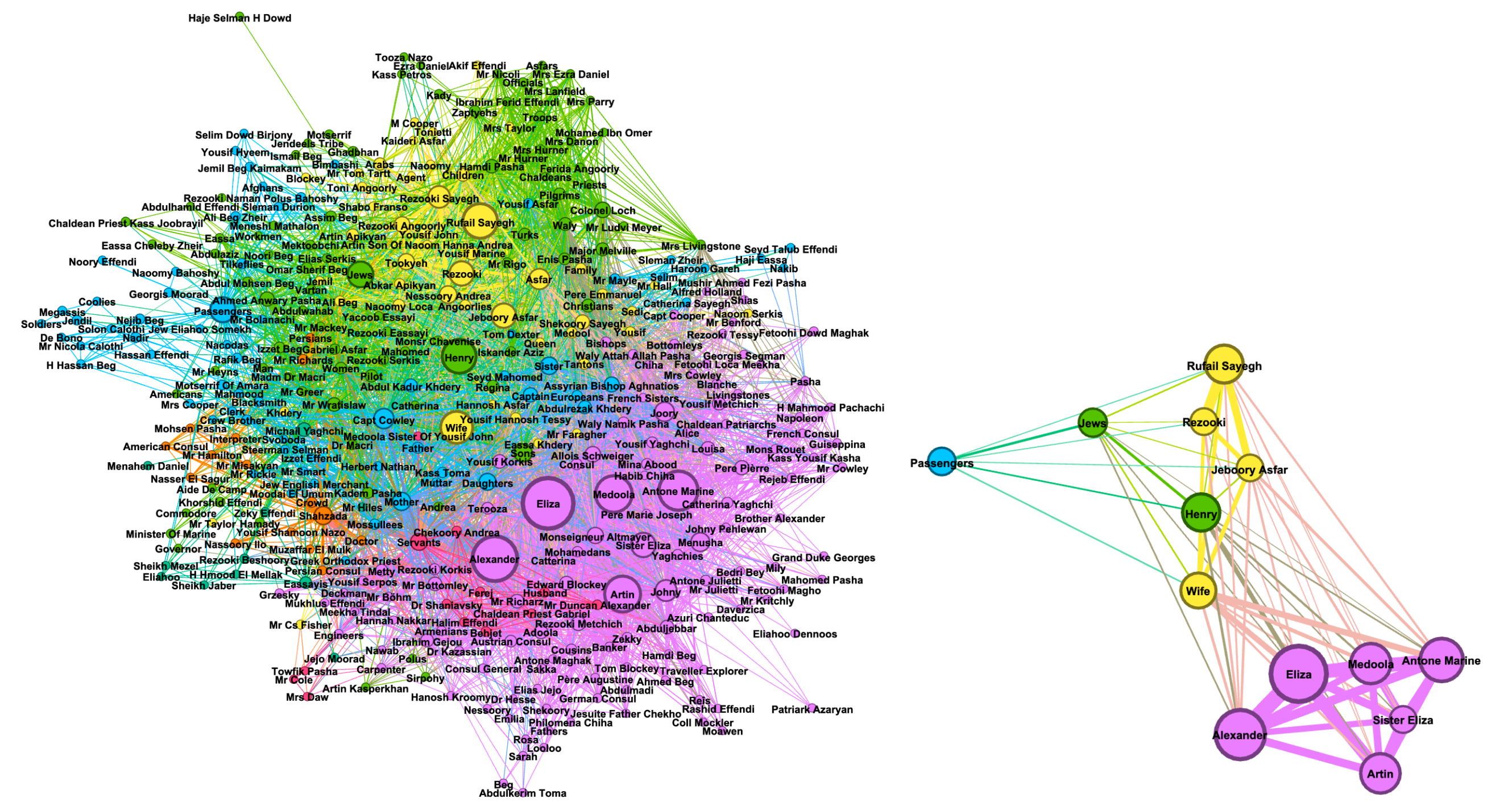

Figure 2: Network visualizations of the persons that Joseph Svoboda mentions in diary 49. Full network (left) vs. only those who appear at least 25 times (right).

Questions of scale arise when we consider the number of persons to visualize. On the one hand, we could focus on a more complete picture of the persons that he mentions in a given diary (Figure 2, left). Or we might focus on the most frequently mentioned persons in a given network (Figure 2, right), which would result in a much simpler visual representation.

At the outset, the network on the right side of Figure 2 is perhaps more "digestible." The simpler network shows the same main modularity groupings as those that appear in the larger network – one that is more focused on Svoboda's social circle in Baghdad (purple), one on his social circle in Basra (yellow), and two depicting people associated with his work on the ships (green and blue). It clearly indicates the most important individuals and groups in each sphere of his life, helping us draw conclusions about the composition of his inner circles.

But we argue that the more complex network on the left facilitates close study in a way that the network on the right does not. The network on the left is not simply a more complete version of the one on the right; the question of scale here is not a simple matter of bigger and smaller. There are a number of reasons for this. First, in the network on the left, we can see minor persons who may be present in Svoboda's life, who play supporting roles to persons who have more frequent or significant interactions with Svoboda. The more comprehensive network affords views of this interrelatedness. Second, the more complex visualization on the left intimates that, though there may primarily be four main groups of individuals that are mentioned in this diary, there may be minor groupings or subplots of interest.

To understand the full gravity of our decisions, we might also consider who to include in the visualization based on who would be left out. Figure 3 depicts the frequency at which different people appear in diary 49. Based on this distribution, we can see that a high proportion of people appear in a single volume of Svoboda's diaries only once. Removing individuals who do not appear often may focus our attention a bit more, but this not only decontextualizes the "main" actors, but potentially also sends an unintentional message about what kinds of interactions and social relations are important. While it can certainly be valid to remove persons from a visualization in order to facilitate analysis by reducing cognitive overload, it is important to be clear about the rationale for reductions in scale, and to not only rely on the "smaller" visualization at the expense of the messier, more contextualized network.

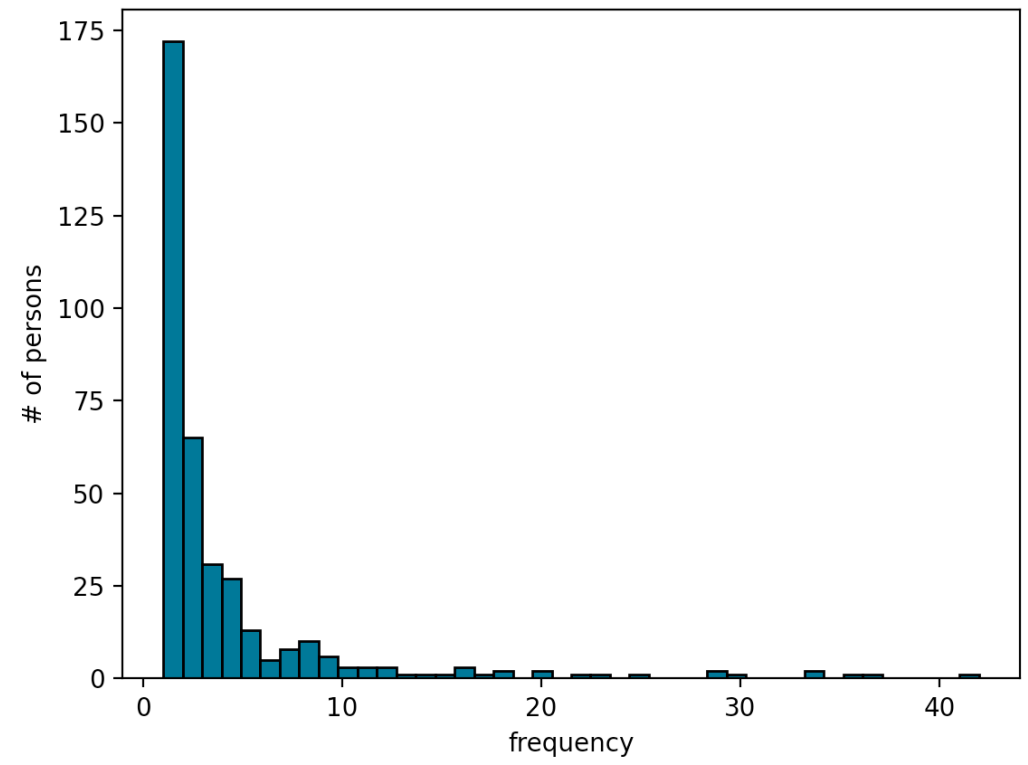

Figure 3: Frequency of Person Mentions by Number of Days. Diary 49.

### 3.2 Representing data

The question of context is also about how both Svoboda and we as researchers represent the members of his social network and milieu. In recent years, there has been increased interest in social factors such as culture, ideology, social norms, and social relations in NLP [24]. Visualizations of textual corpora can facilitate exploration of these concepts. For example, the network that we present in Figure 2 provides Joseph Svoboda's social context in terms of persons involved, but does not illustrate his perceptions of these persons or their routine forms of interaction. We can capture this context by visualizing the different ways that Svoboda refers to these people, in both quantitative and qualitative terms.

The analysis of persons is both informed by extant research questions and emergent from the data. To characterize the persons who appear most frequently in Svoboda's diaries, we manually identified all of the persons as kin or non-kin, where kin were defined as all of Svoboda's blood relations and relations-by-marriage within one degree.

In addition to specific persons, the network visualizations reveal that Joseph often uses descriptors when referring to individuals. At times, he may use a referent that refers to a group more generally ("Jews"); at other times, he uses referents ("Waly of Basreh") that can resolve to several people. He also employs relational terms such as "wife" to describe persons in terms of their relations to others. These "non-specific" nodes were derived through heuristics based on Svoboda's language. For example, in the case of an unnamed person who may be referred to as "wife of X" or "X's wife", that reference would be captured as "wife", with the exception of "my wife" which occurred so frequently it was captured as "Eliza", the name of Svoboda's wife. In cases where unnamed persons, e.g., "Turkish Officer" were mentioned, these were also often resolved to reflect a group characteristic that Svoboda used.

To distinguish between kin, non-kin, and different types of non-specific referents, we created dictionaries. Though our work could facilitate a finer-grained analysis, the depiction of person appearances in Figure 4 enables the analyst to assess the relative prevalence of these different types of referents in Svoboda's writing. From this, we can see that kin constitute the most frequent of Svoboda's interactions, though there is a long tail of persons not depicted. His language is sprinkled with words referring to groups of persons, such as "passengers," "Jews," and "Waly."

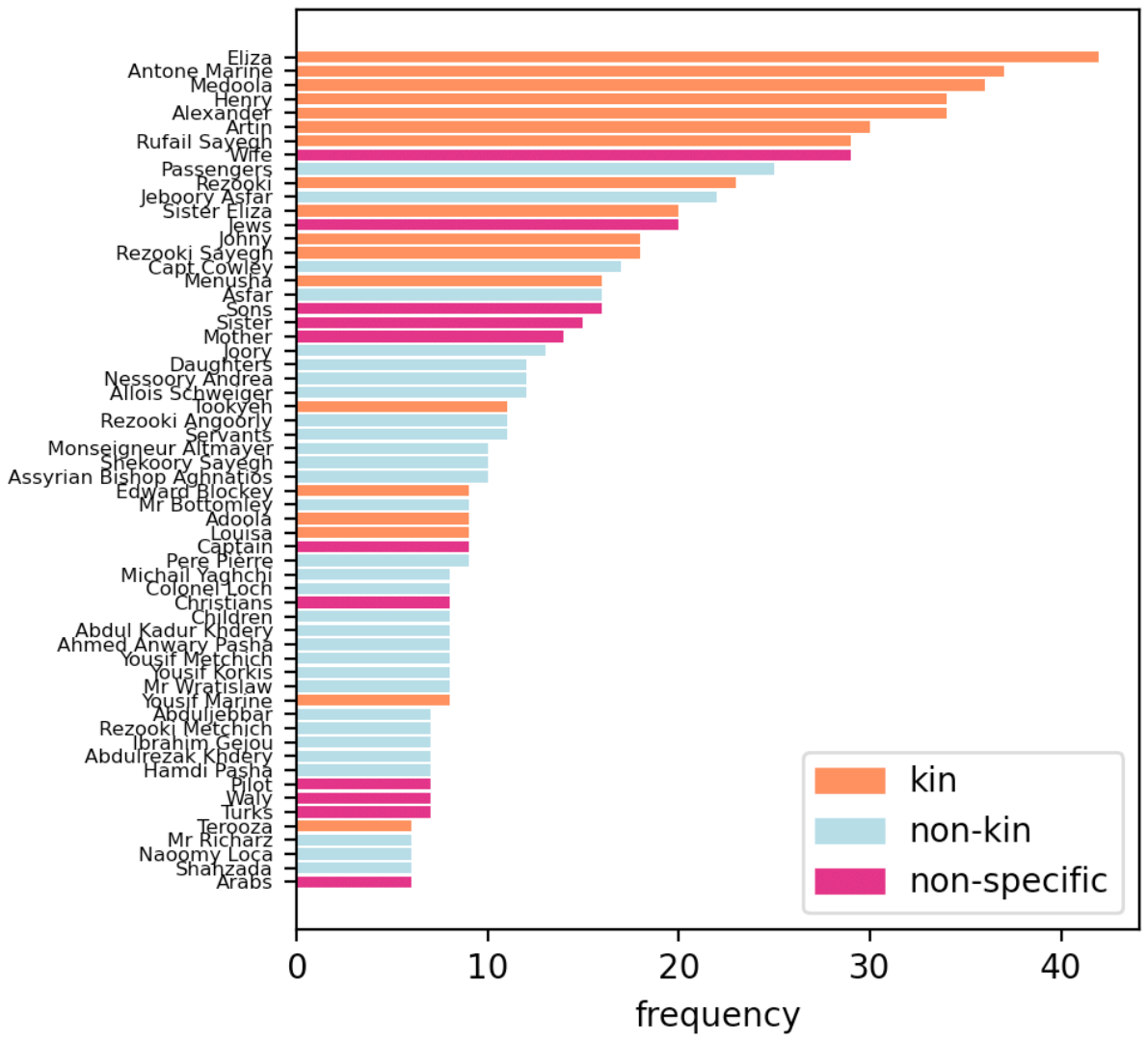

Figure 4. Kin, Non-Kin, and Non-Specific Referents in Diary 49. Persons appearing 5 or more times.

We might ask ourselves how often these descriptors appear, how the same referent is used in conjunction with other descriptors (e.g., "Turkish Pashas"), and in what context. For example, when a descriptor is used, is it used in the context of Svoboda's observations of passengers on the ship, in conversation with others, or in some other manner? Does his tone convey a certain sentiment?

Our current network visualizations include the collapsing of some references to persons into generic categories, e.g., "Turkish Troops, Turkish Pashas, Turkish Man, Turkish Lady, Turkish Officer, Turk, Turkish Employèes," which may afford some level of inference about the "group" (Turks) in relation to the world of Joseph Svoboda on a more holistic level. But it is important to keep in mind that this may result in the collapse of persons who may not occupy the same role in society. The composite terminology of this set of terms reflects concepts concerning gender and occupation. These could also potentially serve as generic categories: for example, it might also make sense to collapse "Turkish Lady" with "Indian Woman," "Mohamedan Woman," and "Woman." Either of these choices reflects a certain evaluation of Svoboda's priorities and the determining characteristics of his social milieu. No choice is perfect; each will render some aspects of the data more visible than others. That choice depends, here and for other research projects, on the categories most salient within the data, and those most relevant to the research questions. More than picking the "right" categorization, however, it is necessary to remain aware of the trade-offs inherent in any such choice; to consider complementary analyses which may offset, explore, or explain gaps or invisibilities in the original approach; and to be as transparent in these choices as possible for downstream users/viewers.

Another place where we have made choices around Svoboda's descriptions involves the range of ways he refers to individuals. For example, Mr. Wratislaw is often referred to in reference to his position "English Consul," and Rejeb Effendi has been referred to as, "Nakib of Basreh Rejeb Effendi son of Seyd Mahdi Said Effendi," indicating both his occupation and his family history. In addition, Svoboda sometimes refers to persons with the title "Mr." and sometimes without it. In network analysis, we generally mapped these referents to a simplified name for each person. However, in doing so, we lose the social nuance of Svoboda's relationship to these persons, and the subtle differentiations of status reflected in how Svoboda uses forms of address.

Previous research on style in literary and historical works has shown that language can provide important information about society and culture. For example, terms of address can carry important social information, including where a person is from and their title and position [25]. The study of kinship terminology in China has facilitated deeper understandings of social interactions among different ethnic groups [26]. In our data, studying the ways that Svoboda refers to different individuals and groups could potentially offer some insight into social interactions of the period. These are not as salient in our current network visualizations, but we aim to study these in future work.

### 3.3 Interactive Visual Analytics in a Team Context

In digital humanities projects today, interdisciplinarity is commonplace [27], but questions remain about how to optimally integrate tools into research [28] and collaborate effectively [29].

The Svoboda Diaries Project is a primarily undergraduate DH lab which aims to provide a flexible yet nurturing environment in which students are able to pursue their own interests as well as acquire the skills and experience that they need to be competitive in whatever career path they choose. The lab is diverse in terms of the umbrella of tasks undertaken, including transcription, software and tool development, computer-assisted historical research, user experience design and development, and dissemination of its products. While this diversity of function can be an enriching experience for all concerned, the structure raises questions about optimal ways to facilitate this learning experience while also ensuring research quality.

These questions also arise within the context of the NLP and visualization pipeline. As indicated in Figure 1 by the person icons,

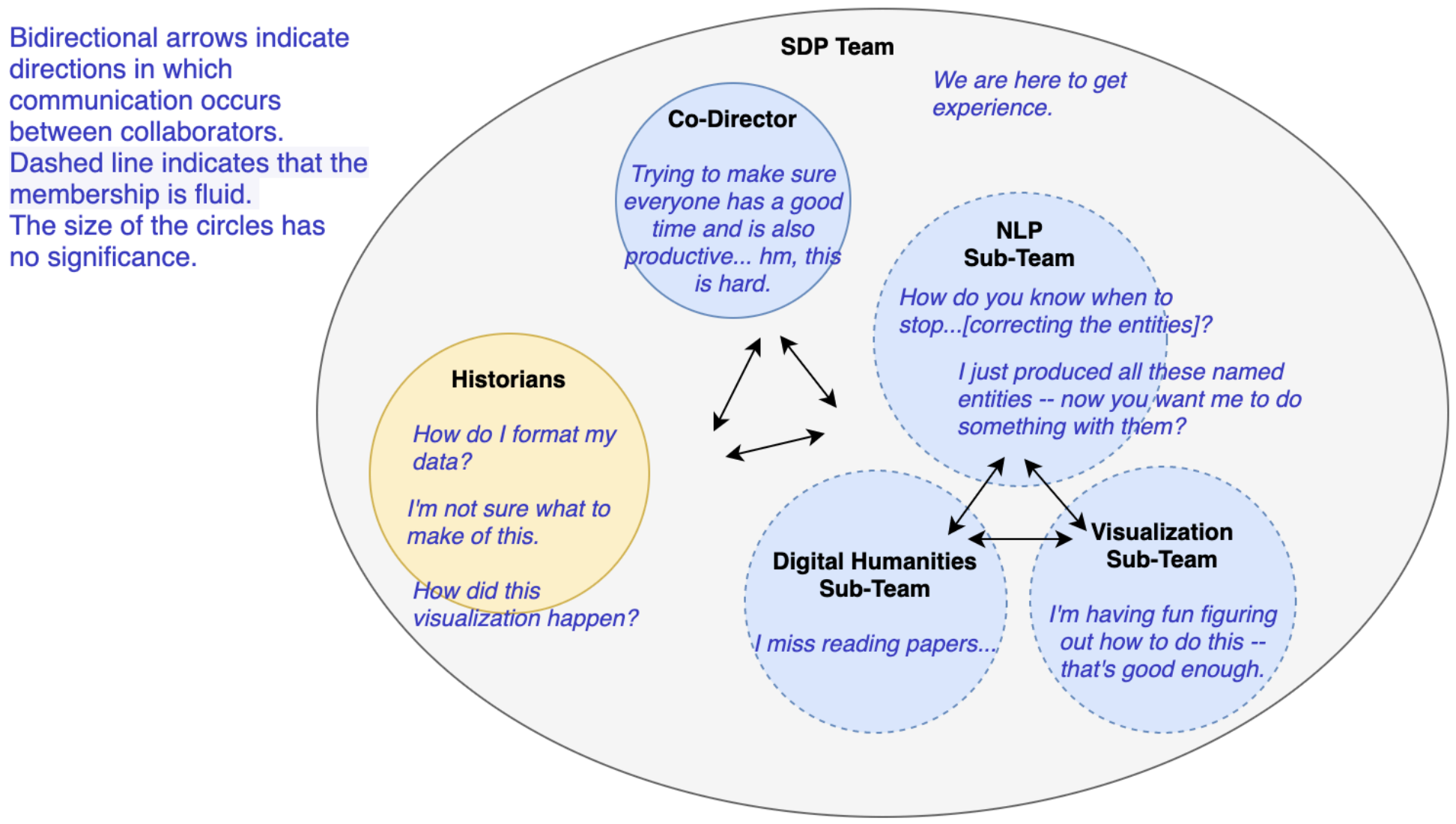

Fig. 5 Cultural Model.

the pipeline involves human interactions at various points. At this juncture, we raise the question of how the visualizations should be read and by whom. Would the same visualization be read similarly by an undergraduate student, a graduate student, or a domain expert? How might discipline influence those readings?

Returning to the left part of Figure 2, interpretation of a high-context figure such as this one requires close study involving moving between the visualization, the 'Data Laboratory' view in Gephi, and the diaries, to form a coherent view of how members of different groups interact or not, and other connections or patterns that may exist. This requires some acquired literacy in how to make sense of figure, spreadsheet, and text in concert. Depending on their training, positionality, and aims, team members may require time as well as different forms of support to develop the literacy necessary to read this kind of figure.

The challenges that arise in acquiring this literacy may have implications for the research process as a whole. First, one issue that can arise with the visualizations is the tendency to use them to support a pre-existing hypothesis. While this may occur with anyone, in the interests of pedagogy as well as research, one idea might be to engage researchers at all different stages of development and background in research sessions to explore the diaries and visualizations interactively.

Extant literature has observed that those who are engaged in DH projects often do not share the same vocabulary or goals [30]. We have also found this to be a challenge within this project. We illustrate this with a cultural model, which is a diagram used in contextual design to show how policy, values, culture, and other factors may affect work outcomes [31], [32] (Figure 5).

This model depicts some of the key actors who work on the NLP, data analysis, and visualization pipeline, who are part of the team. As the figure shows, students bring a lot of energy and enthusiasm to the project, but may also be focused on their more proximate goals, such as acquiring skills in NLP or visualization. Moreover, they often have not had training in the qualitative, social science, and/or historical research methods necessary to engage in interpretation of the text. We provide an outlet for them to learn these skills, but there can be gaps in their understanding of the broader goals of the research.

There are also domain experts (historians) involved in the project, who may communicate with the team regularly as collaborators, or on an ad hoc basis as needs arise, for example, when they use the products of the project. They may bring different types of questions about the project. On one level, the questions may have to do with how to format the data to be able to pursue research questions using visualization; on another, they may have questions about how the visualizations were created or how to employ 'data' or 'visualizations' to answer their research questions.

In Figure 6, we depict a simplified conceptualization of different approaches for starting from text to arrive at an answer to a research question. Whereas one person might work directly from the texts to develop a research question and arrive at an answer, another approach would be to convert the text to data (variables) and/or a visual representation and then answer the research question based on the data (or to use both the text and derived data, as we are currently doing). Some team members may be less focused on research questions, instead concentrating on intermediate steps such as extracting entities (data) from text or generating visualizations, without substantive consideration of subsequent outcomes of the work.

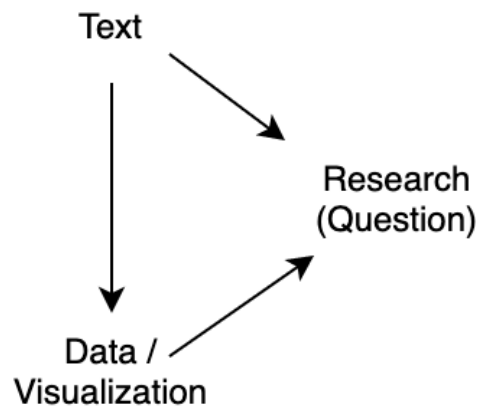

Figure 6. Text-Data-Research Triangle

We argue that these differences in approach have to do with differences in underlying mental models, as well as how each person interacts with research tools in their work. A mental model is a 'small-scale model' of reality which enables people to predict what is going to happen next and decide upon their next steps [33]. How team members conceptualize their work affects how they normally interact with work and research tools. As such, it may take some effort for researchers to learn how to use the tools the development team is using, and for the development team to figure out how to design visualizations which are easily accessible and meaningful for collaborating researchers.

Another challenge is communicating about the entirety of the research process. How a visualization was created has an enormous impact on what interpretations can be derived from it, but either the creator or the interpreter of the visualization may not be aware of this influence or its significance, and essential knowledge about visualizations is not always passed between team members or retained over time. For example, returning to the examples raised in section 3.2, an interpreter who is unaware of how coreference resolution has been performed, or even that it has been performed at all, may draw different conclusions than one who understands the choices made earlier in the pipeline. Conversely, if a creator who is less focused on the research questions makes coreference decisions without documenting or explaining them, it may reverberate down the pipeline. In some cases, creators may not be familiar enough with the corpus to know what resolutions make sense or not.

The question of scale also comes into play here, as the geographic distribution and fluid membership of the research project presents both challenges and opportunities. The fluidity of the group leads to more sharing of ideas and can enhance research productivity. However, that fluidity and geographic spread makes communication and documentation even more critical.

At this point, we return to the question of how to convey the practices involved in generating and interpreting visualizations. As we have shown, knowing how data is extracted, represented, and visualized is critical to interpretation. There are times when this information may not be conveyed between team members due to a lack of awareness of the need to convey it. Things that seem obvious or normal to one person may not be at all obvious to someone with a different background and a different reason for joining the project. We currently record our research process through shared documentation in the cloud and version control. While this is important, if some team members are not attentive to documentation due to a lack of understanding of the importance of choices made in the research process, it can cause problems.

Drucker argues that 'all data are *capta*' (taken and constructed), and though graphic displays often convey the impression of observer-independence and certainty, they are observer co-dependent and interpretive [34]. Increasing awareness of this among team members is particularly important given the potential effect that team members' perspectives on data and representation can have on the research.

Projects and teams are inevitably shaped by their research goals and the actors that constitute the teams. People bring their own interests, motivations, and perspectives into the project; these intersections can yield both intrinsic and extrinsic benefits for the persons involved, as well as contributing to the project as a whole. However, as we have demonstrated, collaboration can also create challenges and barriers, whether due to the differences in epistemic cultures which have been observed between and within disciplines [35], [36], or to differences in expectations and work styles.

Given the diversity of team and research contexts, there is no concrete formula as to how they should be structured. However, our research highlights the need for reflexivity concerning issues of scale and representation throughout the research process, and the need for awareness of potential differences in perspective.

We believe that shared understandings can not only improve research products, but also increase engagement, intellectual enrichment, and felt ownership of the work. To achieve these shared understandings, we suggest communicating with and engaging stakeholders (e.g., team members, external collaborators, and end users of the digital collection) in the shaping of the work. One way in which we have aimed to do this is through regular meetings, which are important for dialogue and rapport. However, meetings alone may not be enough, as some things may best be learned through experience. One solution that we plan to try is to have activity sessions in which we engage team members in tasks involving analysis of the diaries in concert with the visualizations, so that they can better understand how design decisions affect downstream outcomes, as well as how to interact with visualizations. We also suggest employing open-source technologies with version control to enable team members to work together, but care must be taken to convey the importance and utility of documentation and shared understandings so that a shared culture can emerge.

## 4 Discussion

In recent years, alongside interest in big data, there has been an increased awareness of the need for reflexivity and the potential for bias in digital humanities work [7]. In this paper, we presented an example of how the Svoboda Diaries Project has approached questions of scale and representation in our research. In particular, we emphasized flexibility, contextual preservation, and interpretive work. The ways in which we seek to achieve this include the use of open-source technologies, a combination of manual and automated techniques for capturing and representing context, communication, documentation, interdisciplinary collaboration, and reflexivity.

The term "big data" has been associated with the characteristics of volume, variety, velocity, variability, and value [37]. In the context of the digital humanities, Tiepmar has argued that despite the relatively small footprint of textual data, digital humanities research can be considered big data for a number of reasons, including the addition of derived secondary data such as part-of-speech tags, named entities, editorial notes, topic models, co-occurrence and collocation data, and more [38].

In the context of this project, much of our work involves a dataset that may be seen as limited in many ways, including its volume, origination with (mostly) a single author, and single genre/source type. However, the idea of "smallness" does not accurately reflect the issues of scale that we face internal to our analysis, nor does it evoke the "large" questions of social context and categorization faced by Svoboda and by the research team. In this paper, we have illustrated how the size of a dataset, the scope of a project, and the boundaries of a team can be fluid and permeable at any given point and over time. We argue that rather than characterizing data as "big" or "small," or analysis as "micro" or "macro," we should think about how sources, and analyses, can be both simultaneously. By emphasizing reflexivity about the value of different analytic scales, and how they can complement each other, we show that scale is not a fixed or intrinsic property.

In addition to questions of scale, this paper emphasizes the importance of reflexivity at multiple levels and multiple points in the research process. Working with a fluid, multi-sited team which incorporates members with various backgrounds and interests means that reflexivity and attention to the constructed nature of data is both an individual task and one that requires cross-team discussion. We outline the importance of reflexivity at multiple points in the data-creation and representation process, as well as in the course of analysis. The issues that arise in analyzing our visualizations cannot be separated from the team context in which the data are "taken and constructed" [34]. As a result, we argue that questions about representation and scale that arise in analysis should be approached both from the perspective of the research question and considering the context of the research team.

## Acknowledgments

The authors would like to thank Adeline Perkins, Sam Fields, and Catherine Oei for their feedback and contributions to the figures in this paper.